\documentclass[11pt,a4paper]{article}
\usepackage[utf8]{inputenc}
\usepackage{amsmath,amssymb}
\usepackage{graphicx}
\usepackage{hyperref}
\usepackage{booktabs}
\usepackage{geometry}
\usepackage[section]{placeins}

\title{Short-Horizon Predictive Maintenance of Industrial Pumps Using Time-Series Features and Machine Learning}

\author{
Khaled M. A. Alghtus$^{1}$\thanks{Corresponding author: \texttt{khaled@physik.hu-berlin.de}} \and
Aiyad Gannan$^{2}$ \and
Khalid M. Alhajri$^{2}$ \and
Ali L. A. Al Jubouri$^{2}$ \and
Hassan A. I. Al-Janahi$^{2}$
}

\date{
$^{1}$ Humboldt-Universität zu Berlin, Institut für Physik, AG Moderne Optik, Berlin, Germany \\
$^{2}$ University of Doha for Science and Technology, College of Engineering and Technology, 
Department of Mechanical Engineering Technology, Doha, Qatar
}

\begin{document}

\maketitle


\begin{abstract} This study presents a machine learning framework for forecasting short-term faults in industrial centrifugal pumps using real-time sensor data. The approach aims to predict {EarlyWarning} conditions 5, 15, and 30 minutes in advance based on patterns extracted from historical operation. Two lookback periods, 60 minutes and 120 minutes, were evaluated using a sliding window approach. For each window, statistical features including mean, standard deviation, minimum, maximum, and linear trend were extracted, and class imbalance was addressed using the SMOTE algorithm. Random Forest and XGBoost classifiers were trained and tested on the labeled dataset. Results show that the Random Forest model achieved the best short-term forecasting performance with a 60-minute window, reaching recall scores of 69.2\% at 5 minutes, 64.9\% at 15 minutes, and 48.6\% at 30 minutes. With a 120-minute window, the Random Forest model achieved 57.6\% recall at 5 minutes, and improved predictive accuracy of 65.6\% at both 15 and 30 minutes. XGBoost displayed similar but slightly lower performance. These findings highlight that optimal history length depends on the prediction horizon, and that different fault patterns may evolve at different timescales. The proposed method offers an interpretable and scalable solution for integrating predictive maintenance into real-time industrial monitoring systems.
\end{abstract}

\vspace{0.5em}

\section{Introduction}

Centrifugal pumps are essential in petrochemical plants, power stations, and water networks. Failures in these systems disrupt production, reduce efficiency, and create safety risks \cite{ref-hallaji2022}. Maintenance strategies are therefore shifting from reactive or schedule-based approaches to predictive ones. Predictive maintenance (PdM) uses sensor data, signal analysis, and machine learning to anticipate faults before they occur \cite{ref-zonta2020}.  

In rotating equipment, both feature engineering and deep learning methods have shown value. Key signals include vibration, temperature, pressure, and electrical current \cite{ref-mushtaq2021,ref-achouch2022}. Reviews highlight that combining multiple sensor sources with reliable labeling is critical for accurate diagnosis \cite{ref-hallaji2022,ref-meitz2025}.  

Recent work also explores digital twins. These models allow monitoring systems to remain interpretable and update as conditions change \cite{ref-aivaliotis2019}. Researchers continue to debate between using hand-crafted features and end-to-end neural networks. Studies confirm that feature-based methods remain competitive, especially when datasets are small or imbalanced \cite{ref-sanchez2024}. Ensemble and hybrid models have been introduced to improve robustness \cite{ref-li2025,ref-varalakshmi2025}. Physics-informed learning is also gaining attention for improving generalization \cite{ref-ni2023}.  

Most predictive maintenance studies focus on long-term prognostics such as remaining useful life estimation or fault detection. This study instead addresses short-horizon forecasting. The goal is to predict abnormal pump behavior minutes in advance rather than hours or days. Case studies show that such near-term forecasting is valuable because it allows operators to act before a small fault becomes a critical shutdown \cite{ref-pinciroli2024,Zhang2024Electronics}.  

The analysis in this work is based on real sensor data from an industrial facility operating centrifugal pumps under continuous service. The data come from a large petrochemical company in Qatar. Due to confidentiality agreements, the company’s name cannot be disclosed. This ensures the results reflect real-world conditions while maintaining data protection.  

The framework developed here combines fixed thresholds with adaptive statistical limits to create explainable labels. A sliding-window feature extraction method captures mean values, variability, and temporal trends in sensor data. Two history lengths (60 and 120 minutes) and three forecast horizons (5, 15, and 30 minutes) are compared. This design makes it possible to assess how temporal context influences prediction accuracy. The outcome is a practical and interpretable approach for embedding real-time short-horizon fault prediction into pump monitoring systems.  


\section{Methodology}

This study introduces a machine learning framework for short-term fault prediction in centrifugal pumps. The aim is not only to identify the current pump condition but also to forecast whether an abnormal state will occur in the next 5, 15, or 30 minutes. The workflow is based on historical sensor records from Pump P-65401A and consists of data preparation, dual-threshold labeling, sliding window feature extraction, data balancing, and supervised model training.

\begin{figure}[htbp]
    \centering
    \includegraphics[width=0.85\textwidth]{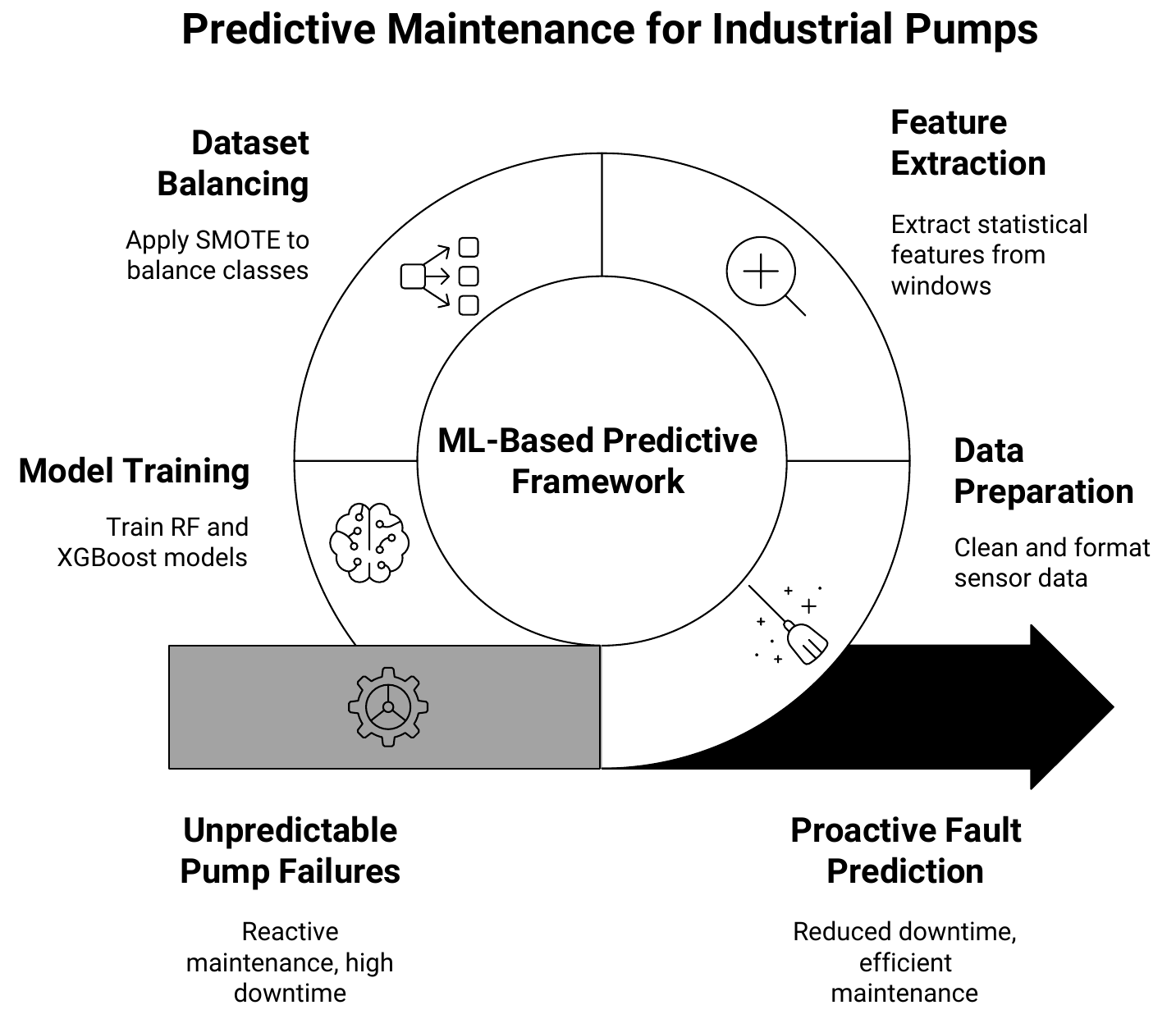}
    \caption{Workflow of the predictive maintenance method, starting from raw sensor data and progressing through preprocessing, labeling, sliding window feature extraction, SMOTE balancing, and model training for future condition prediction.}
    \label{fig:workflow}
\end{figure}

\subsection{Data Source and Preparation}

The dataset was collected from an industrial facility, where the monitored pump continuously logged five signals at one-minute intervals: vibration (mm/s), temperature (°C), flow (m$^3$/h), pressure (bar), and electrical current (A). Each record contained a timestamp and the corresponding sensor readings. Data preparation involved converting timestamps into a standard datetime format, removing invalid entries, and filling small gaps to obtain a continuous multivariate time series. This process ensured that the dataset was ready for feature extraction and supervised learning.

\subsection{Labeling with Fixed and Adaptive Thresholds}

A dual-threshold labeling scheme was applied to classify the operational state. The fixed threshold represents an engineering limit, while the adaptive threshold corresponds to the 95th percentile of historical sensor values. Observations above the fixed threshold were labeled as \textit{CriticalAlert}, values between adaptive and fixed thresholds were labeled as \textit{EarlyWarning}, and the rest as \textit{Normal}. This approach captures both safety-critical exceedances and statistically unusual deviations, and is consistent with earlier predictive maintenance studies in rotating equipment \cite{ref-li2023,ref-zonta2020}. At each time step, the overall pump condition was assigned according to the most severe sensor label.

\subsection{Initial Attempt with LSTM Modeling}

An initial attempt used a Long Short-Term Memory (LSTM) network trained on raw sequences of 60 minutes of data. The goal was to predict the future pump condition at different horizons. Although LSTMs are designed for sequence modeling, training was unstable and performance was weak. Similar challenges with deep learning under imbalanced or limited industrial datasets have been noted in previous reviews \cite{ref-achouch2022}. These observations motivated a shift toward statistical feature engineering, which provides interpretable and robust descriptors of recent system behavior.

\subsection{Sliding Window Feature Extraction}

To capture temporal patterns, a sliding window method was applied. Two window lengths were tested: 60 minutes and 120 minutes. Each window was shifted forward by one minute, creating overlapping samples. For each sensor within a window, five features were calculated: mean, standard deviation, minimum, maximum, and linear trend (slope). This produced a feature vector of 25 values. The label was defined by the pump condition at the forecast horizon. Such sliding-window feature extraction has been shown to balance interpretability with accuracy in predictive maintenance applications \cite{ref-sanchez2024,ref-pinciroli2024}.

\begin{figure}[htbp]
    \centering
    \includegraphics[width=0.71\textwidth, angle=-90]{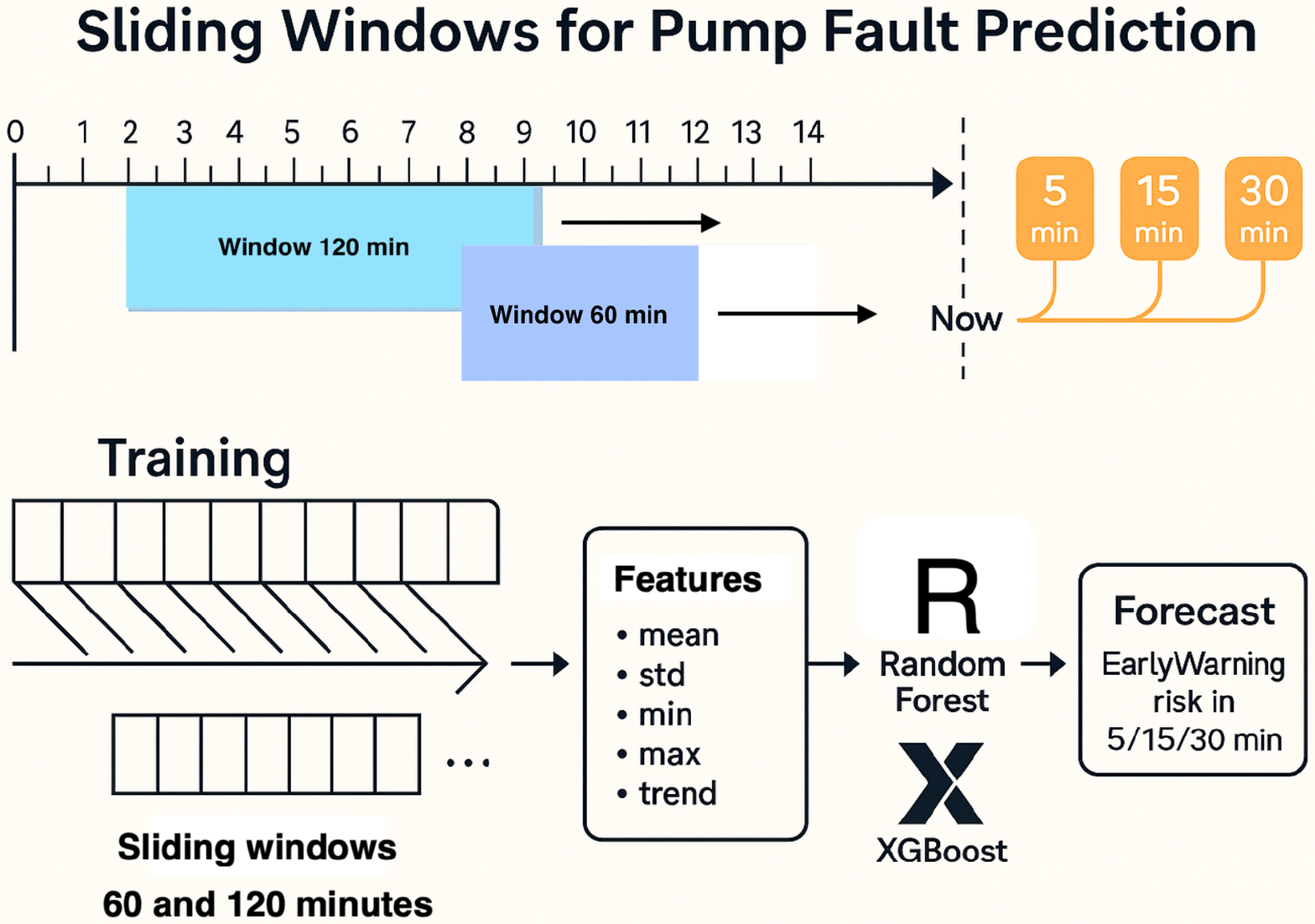}
    \caption{Illustration of the sliding window mechanism. Historical windows of 60 or 120 minutes are used to predict pump condition 5, 15, or 30 minutes into the future. Each window slides by one minute.}
    \label{fig:sliding_window}
\end{figure}

\subsection{Dataset Balancing with SMOTE}
The dataset showed strong class imbalance, with many more \textit{Normal} samples than \textit{EarlyWarning} or \textit{CriticalAlert}. This is a well-known issue in predictive maintenance and time-series classification \cite{Zhu2022KBS,Jiang2019IEEEAccess}. To address this, the Synthetic Minority Oversampling Technique (SMOTE) was applied to the training set. SMOTE generates synthetic samples of minority classes by interpolating between existing data points. Prior studies confirm that such resampling methods improve predictive performance in industrial scenarios \cite{ref-mahale2025}. This step ensured that classifiers were trained on more balanced data.

\subsection{Model Training and Evaluation}

Two supervised algorithms were selected: Random Forest and XGBoost. Each was trained under six configurations, combining two input window lengths with three forecast horizons. Models were trained on 75\% of the data and tested on the remaining 25\%. The primary evaluation metric was recall for the \textit{EarlyWarning} class, since missed detections pose greater operational risks than false alarms. Confusion matrices were also used to analyze error patterns. Ensemble classifiers such as Random Forest and XGBoost have been widely used in industrial condition monitoring due to their robustness and competitive accuracy \cite{ref-almazrouei2024,ref-varalakshmi2025}.

\subsection{Evaluation Metrics and Statistical Testing}

Model performance was assessed using recall, precision, F1-score, false alarm rate (FAR), and area under the ROC curve (AUROC). Confidence intervals at the 95\% level were estimated using bootstrap resampling. To evaluate whether differences between models were significant, McNemar’s test was applied to paired classification outcomes.

\subsection{Baselines}

For context, model performance was compared against simple baselines:  
(i) fixed-threshold rule,  
(ii) adaptive-threshold rule,  
(iii) persistence (predicting no change),  
(iv) majority class (always predicting Normal),  
(v) logistic regression on the same features, and  
(vi) an optional Isolation Forest trained on Normal samples.  

\subsection{Ablation Studies}

Additional experiments were conducted to isolate the contribution of individual design elements. Variants included removing SMOTE, limiting features to mean and standard deviation, restricting sensor subsets, and testing simplified labeling schemes. Window lengths (60 vs.\ 120 minutes) and horizons (5, 15, 30 minutes) were also varied. Such ablation studies are common in predictive maintenance research to quantify the effect of design choices \cite{ref-meitz2025,Zhang2024Electronics}.

\subsection{Notation and Key Formulas}

For consistency between the data analysis implementation and the predictive maintenance framework, the following notations and key formulas are introduced.

\paragraph{Sensor Variables:}  
The pump condition was monitored using five sensor signals:
\begin{equation}
x(t) = \{ V(t),\, T(t),\, F(t),\, P(t),\, C(t) \},
\label{eq:sensors}
\end{equation}
where  
$V(t)$ = vibration (mm/s),  
$T(t)$ = temperature (°C),  
$F(t)$ = flow rate (m$^{3}$/h),  
$P(t)$ = pressure (bar),  
$C(t)$ = electrical current (A).  

\paragraph{Thresholds:}  
For each signal $s(t)$, two thresholds were defined:  
\begin{itemize}
    \item Fixed engineering threshold: $T^{\text{fixed}}_{s}$ (from datasheets and field standards),  
    \item Adaptive statistical threshold: $T^{95\%}_{s}$, the 95th percentile of the historical distribution.  
\end{itemize}

\paragraph{Labeling Rule:}  
Each sample was assigned a label $y(t)$ according to:
\begin{equation}
y(t) =
\begin{cases}
\text{CriticalAlert}, & s(t) > T^{\text{fixed}}_{s}, \\
\text{EarlyWarning}, & T^{\text{fixed}}_{s} \geq s(t) > T^{95\%}_{s}, \\
\text{Normal}, & s(t) \leq T^{95\%}_{s}.
\end{cases}
\label{eq:labeling}
\end{equation}

\paragraph{Sliding Window Feature Extraction:}  
For a sequence length $L$ (either 60 or 120 minutes), statistical features were computed:
\begin{equation}
f = \{\mu,\, \sigma,\, \min,\, \max,\, \text{trend}\},
\label{eq:features}
\end{equation}
with
\begin{equation}
\mu = \frac{1}{L}\sum_{i=1}^{L} s_i, 
\quad 
\sigma = \sqrt{\frac{1}{L}\sum_{i=1}^{L}(s_i-\mu)^2},
\quad 
\text{trend} = \text{slope of linear fit}.
\label{eq:stats}
\end{equation}

This produced $5 \times 5 = 25$ features for each window across the five sensors.

\paragraph{Model Training:}  
The classification task was to predict $y(t+\Delta)$ at forecast horizons $\Delta \in \{5, 15, 30\}$ minutes, using input features from the preceding $L$ minutes.  

Random Forest (RF) prediction rule:
\begin{equation}
\hat{y} = \text{mode}\{h_{1}(x), h_{2}(x), \dots, h_{T}(x)\},
\label{eq:rf}
\end{equation}
where $h_{i}$ is the $i$-th decision tree.  

XGBoost optimization objective:
\begin{equation}
\mathcal{L} = \sum_{i=1}^{n} \ell(y_i, \hat{y}_i^{(m)}) + \sum_{t=1}^{T}\Omega(h_t),
\label{eq:xgb}
\end{equation}
where $\ell$ is the loss function and $\Omega$ is a regularization penalty.  

\paragraph{Evaluation Metrics:}  
Performance was reported using recall, precision, F1-score, false alarm rate, and AUROC. Since missed detections are more costly, recall for the \textit{EarlyWarning} class was the primary focus.

\section{Results}

This section presents the outcomes of the time-series forecasting framework. The analysis begins by detailing the data labeling process based on the dual-threshold strategy. Subsequently, the performance of the machine learning models in predicting future `EarlyWarning` events is evaluated across multiple time horizons. The section concludes with an analysis of the most important features that contribute to the models' predictive capabilities.

\subsection{Sensor Data Labeling and Anomaly Identification}

The condition of the pump was monitored using five key operational parameters: vibration (mm/s), temperature (°C), flow rate (m$^3$/h), pressure (bar), and electrical current (A). Each of these sensor signals was recorded continuously under real-world operational conditions. To detect anomalies and precursors to failure, each time series was evaluated using two thresholding strategies: fixed critical thresholds and adaptive early warning thresholds.

The fixed thresholds represent absolute engineering limits, typically based on manufacturer specifications or field engineering standards. These thresholds define the boundary beyond which a measurement indicates an imminent or critical fault. In contrast, the adaptive thresholds were calculated from the actual sensor data using the 95th percentile of historical values. This approach captures unusual or degrading behavior that may not have reached the critical zone but still reflects a departure from normal operation.

Table~\ref{tab:thresholds} summarizes the thresholds used for all five parameters. The fixed thresholds were used to identify critical alerts, while the adaptive values served to trigger early warnings.

\begin{table}[htbp]
    \centering
    \caption{Summary of thresholds used for data labeling, combining fixed engineering limits and adaptive 95th percentile values.}
    \label{tab:thresholds}
    \begin{tabular}{lcc}
        \toprule
        \textbf{Sensor Parameter} & \textbf{Critical (Fixed)} & \textbf{Early Warning (Adaptive)} \\
        \midrule
        Vibration (mm/s)       & 5.00                      & 1.65                              \\
        Temperature (°C)       & 80.00                     & 55.23                             \\
        Flow (m³/h)            & 2800.00                   & 2668.05                           \\
        Pressure (bar)         & 6.00                      & 4.77                              \\
        Current (A)            & 240.00                    & 231.89                            \\
        \bottomrule
    \end{tabular}
\end{table}

\begin{figure}[htbp]
\centering
\includegraphics[width=0.85\textwidth]{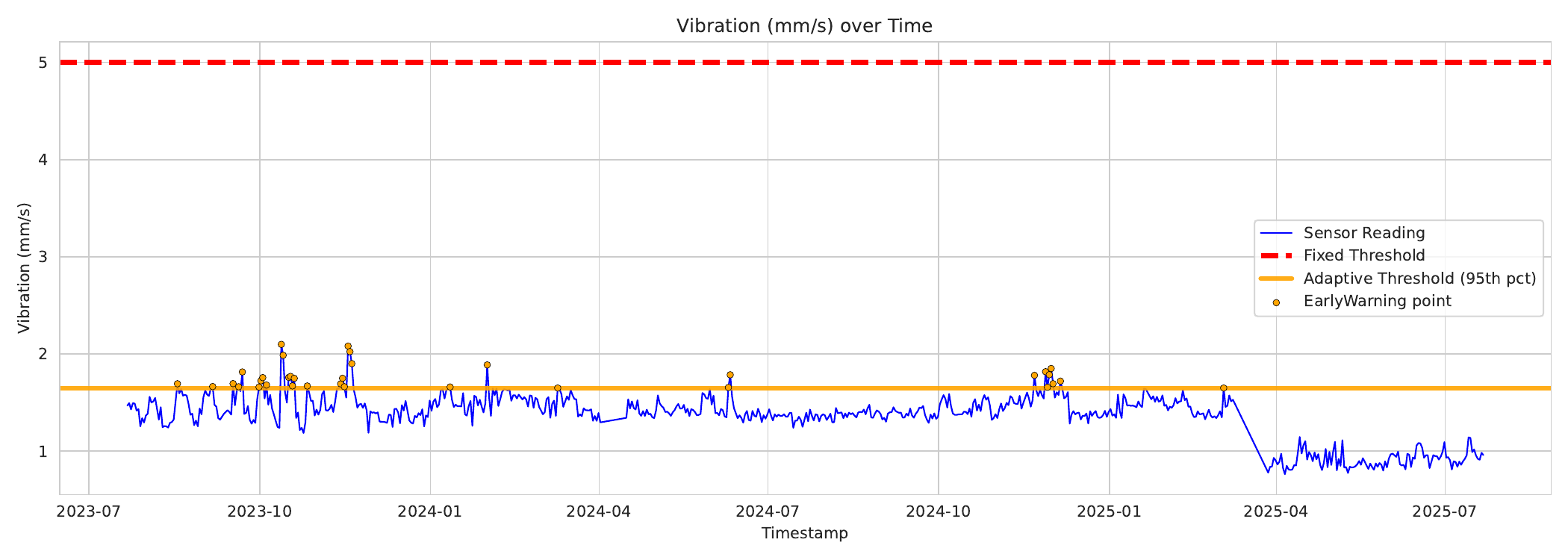}
\includegraphics[width=0.85\textwidth]{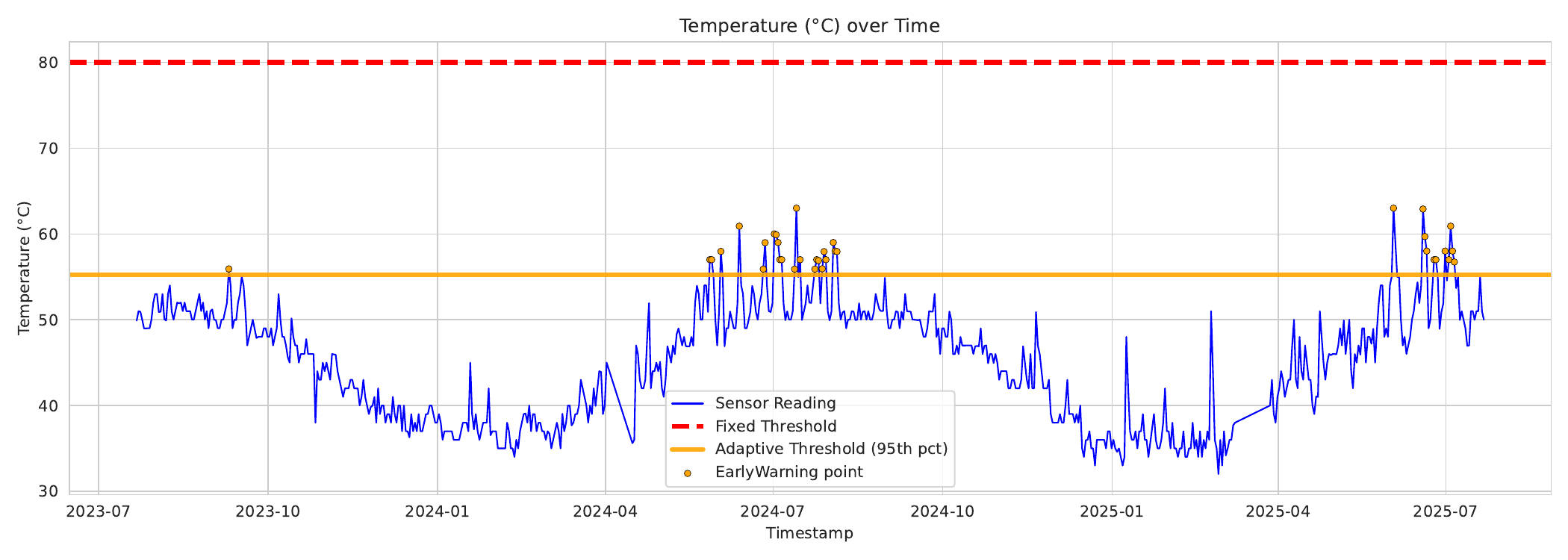}
\includegraphics[width=0.85\textwidth]{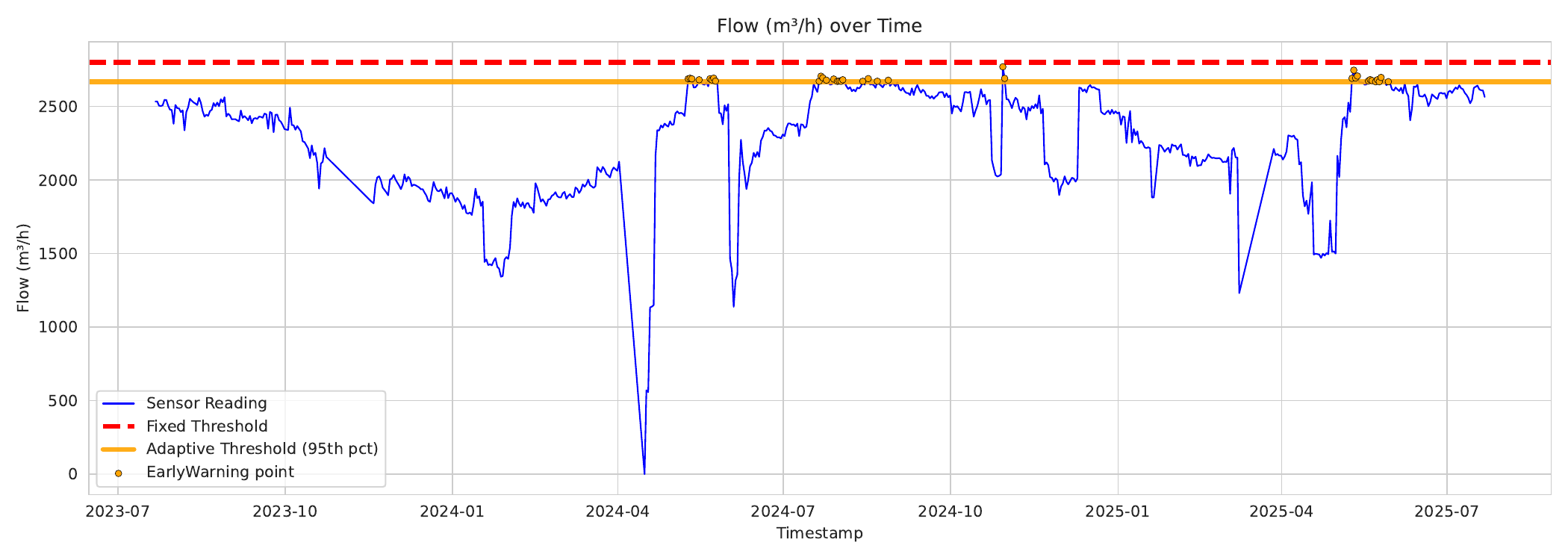}
\includegraphics[width=0.85\textwidth]{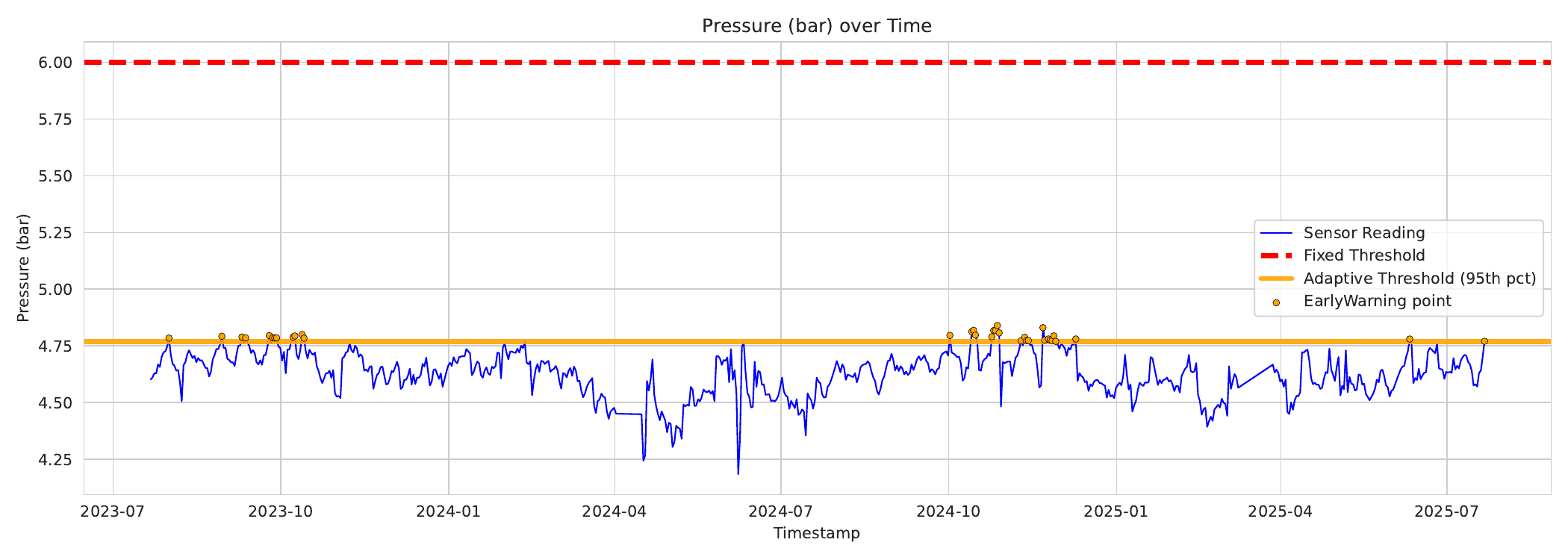}
\includegraphics[width=0.85\textwidth]{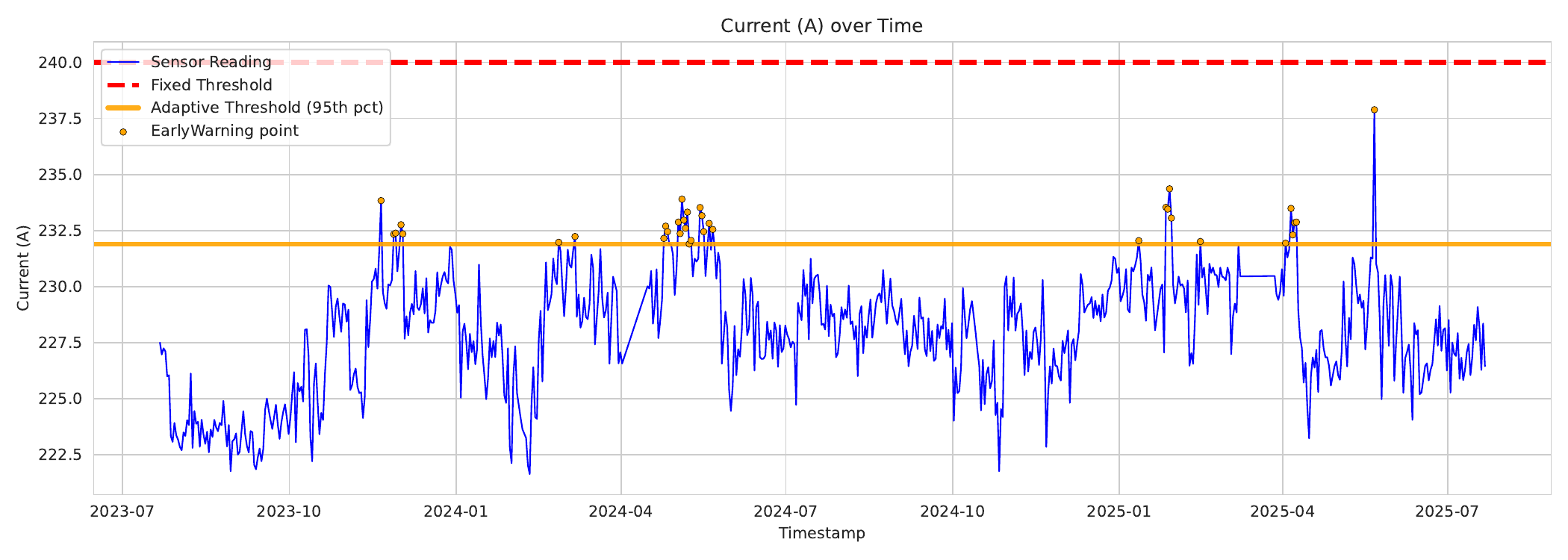}
\caption{Sensor monitoring plots with both fixed (red dashed) and adaptive (orange dotted) thresholds. Orange points indicate early warning events.}
\label{fig:thresholds}
\end{figure}

Figure~\ref{fig:thresholds} presents the complete time series for each sensor parameter, overlaid with the corresponding thresholds. Red dashed lines indicate the fixed limits, and orange dotted lines represent the adaptive thresholds. Orange markers highlight individual data points that exceed the adaptive threshold but remain below the critical level, thus classified as early warnings.

The vibration signal stayed well below the critical threshold, with a number of short-duration deviations triggering early warnings, often during localized load increases. The temperature profile showed long-term thermal fluctuations, including periodic clusters of early warning events, especially during warmer operational periods. The flow rate signal exhibited behavior typical of demand-regulated systems, with most values close to the upper range, which explains why early warnings appeared relatively near the fixed threshold. Pressure and current readings were mostly stable, with intermittent spikes that briefly crossed the adaptive boundary, resulting in isolated early warnings. These thresholds enabled a consistent, explainable classification of operational states and served as the foundation for machine learning model training in later stages.

\subsection{Predictive Model Performance}

The trained machine learning models were evaluated using two different sliding window lengths: 60 minutes and 120 minutes. Each configuration aimed to predict the pump's condition at future time horizons of 5, 15, and 30 minutes. The prediction task involved classifying the upcoming pump state as either Normal or EarlyWarning, based on patterns observed in the past time window.

\begin{figure}[htbp]
    \centering
    \includegraphics[width=0.99\textwidth]{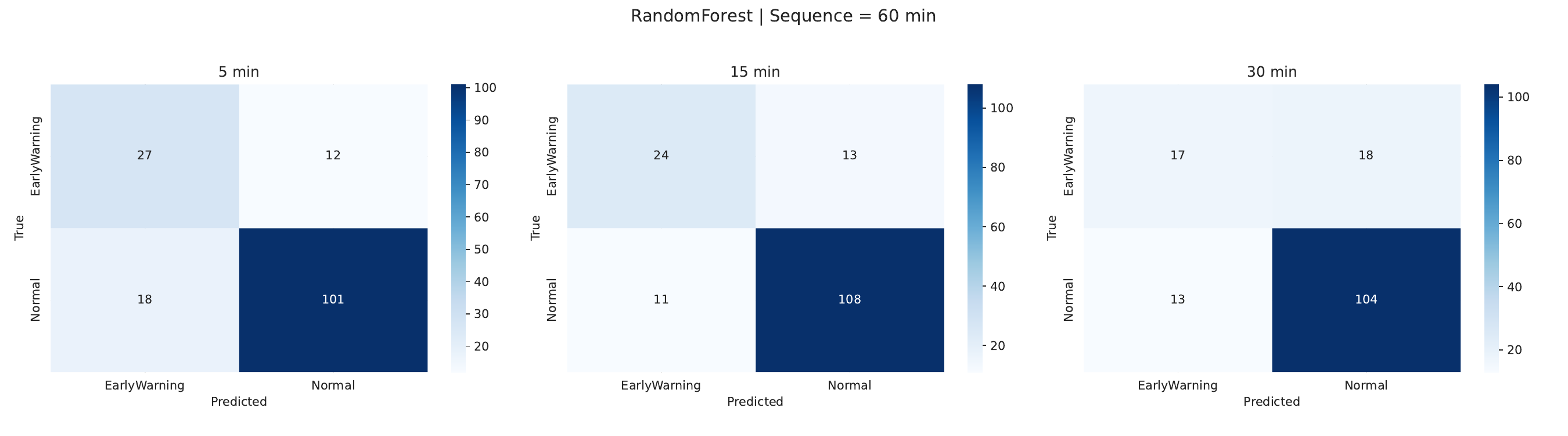}
    \includegraphics[width=0.99\textwidth]{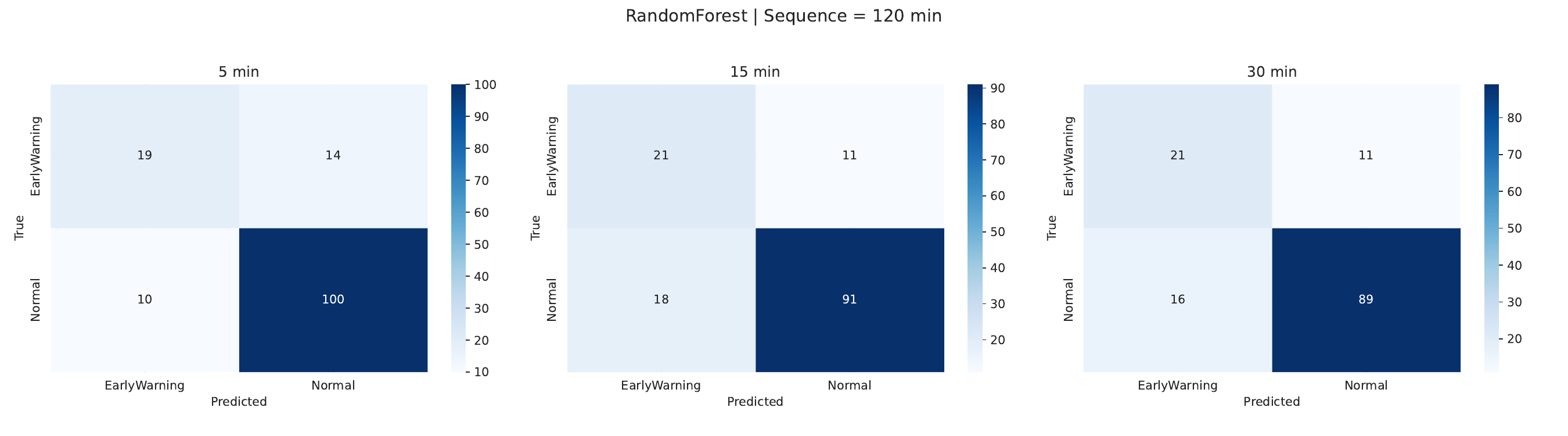}
    \caption{Confusion matrices for the Random Forest classifier at 60-minute and 120-minute sequence lengths. Each matrix shows the number of correct and incorrect predictions for EarlyWarning and Normal classes at 5, 15, and 30-minute horizons.}
    \label{fig:confusion_rf}
\end{figure}

\begin{figure}[htbp]
    \centering
    \includegraphics[width=0.99\textwidth]{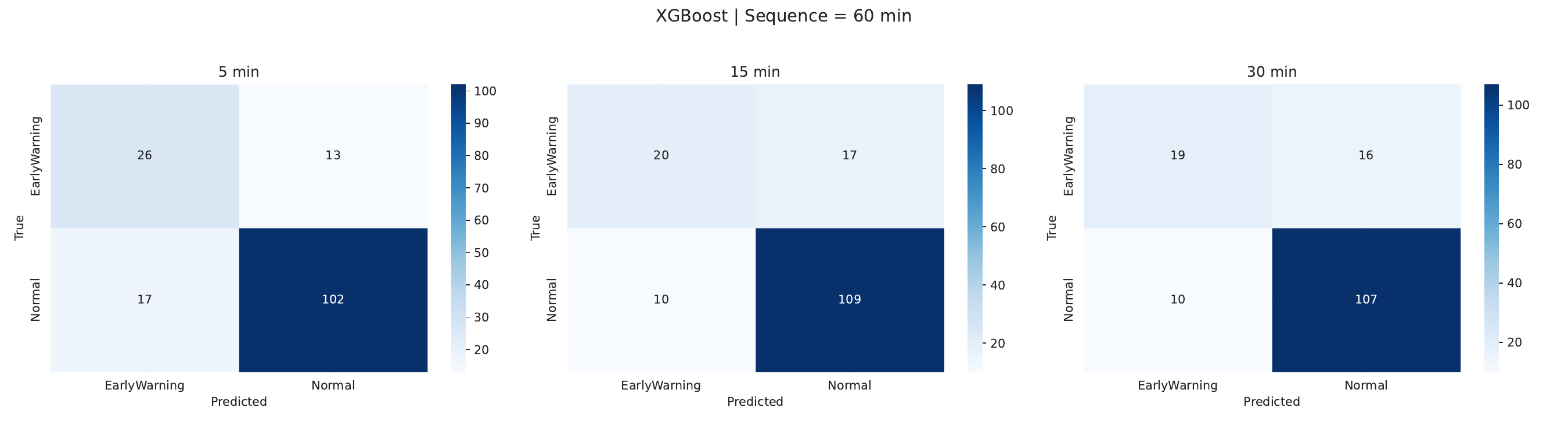}
    \includegraphics[width=0.99\textwidth]{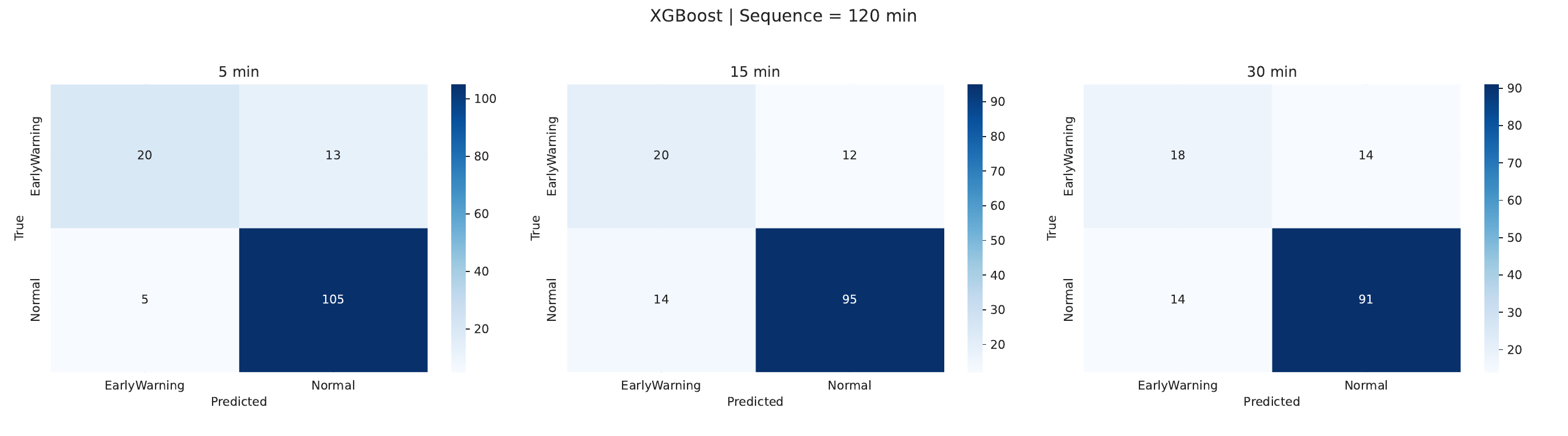}
    \caption{Confusion matrices for the XGBoost classifier at 60-minute and 120-minute sequence lengths. Each matrix shows the number of correct and incorrect predictions for EarlyWarning and Normal classes at 5, 15, and 30-minute horizons.}
    \label{fig:confusion_xgb}
\end{figure}

Figure~\ref{fig:confusion_rf} and Figure~\ref{fig:confusion_xgb} present the confusion matrices for the Random Forest and XGBoost classifiers, respectively. Each matrix provides a breakdown of model predictions versus actual labels. In each square grid, the top-left cell represents true EarlyWarning cases correctly identified, the bottom-right shows correct Normal predictions, while the off-diagonal cells correspond to misclassifications. This structure helps quantify the model's strengths and limitations.

For example, in the Random Forest model with a 60-minute window and a 5-minute horizon, the confusion matrix shows that out of 39 true EarlyWarning cases, 27 were correctly predicted while 12 were missed. Similarly, out of 119 true Normal cases, 101 were correctly classified and 18 were incorrectly labeled as EarlyWarning. As the prediction horizon increases to 30 minutes, the number of missed early warnings increases, indicating that forecasting further into the future becomes more difficult.

Using the 120-minute sequence window, some performance shifts were observed. For Random Forest at the 30-minute horizon, 21 out of 32 true EarlyWarning cases were correctly detected, and 89 out of 105 Normal instances were also correctly predicted. These results suggest that a longer historical context improves sensitivity to upcoming faults, particularly at longer forecast horizons.

Table~\ref{tab:recall_summary} summarizes the recall scores for each model and time horizon. The EarlyWarning recall reflects how well the model identifies potential issues in advance, while the Normal recall measures its reliability in recognizing healthy operation. Recall values closer to 1.0 indicate better performance.

\begin{table}[htbp]
    \centering
    \caption{Summary of model recall scores for predicting EarlyWarning and Normal states across the 5, 15, and 30-minute time horizons.}
    \label{tab:recall_summary}
    \resizebox{\textwidth}{!}{%
    \begin{tabular}{lcccccc}
        \toprule
        & \multicolumn{2}{c}{\textbf{5 min Horizon}} & \multicolumn{2}{c}{\textbf{15 min Horizon}} & \multicolumn{2}{c}{\textbf{30 min Horizon}} \\
        \cmidrule(lr){2-3} \cmidrule(lr){4-5} \cmidrule(lr){6-7}
        \textbf{Model} & \textbf{EarlyWarning} & \textbf{Normal} & \textbf{EarlyWarning} & \textbf{Normal} & \textbf{EarlyWarning} & \textbf{Normal} \\
        \midrule
        RandomForest (60 min) & 0.692 & 0.849 & 0.649 & 0.908 & 0.486 & 0.889 \\
        XGBoost      (60 min) & 0.667 & 0.857 & 0.541 & 0.916 & 0.543 & 0.915 \\
        RandomForest (120 min) & 0.576 & 0.909 & 0.656 & 0.835 & 0.656 & 0.848 \\
        XGBoost      (120 min) & 0.606 & 0.955 & 0.625 & 0.872 & 0.562 & 0.867 \\
        \bottomrule
    \end{tabular}
    }
\end{table}

Overall, models trained on 60-minute windows performed slightly better for short-term predictions, offering faster response to developing issues. In contrast, models trained on 120-minute sequences demonstrated improved performance for longer horizons, suggesting that a longer observation window helps capture patterns leading to faults that develop gradually over time. The normal class was consistently well identified, which indicates high stability in model predictions under non-faulty conditions.

\subsection{Analysis of Predictive Features}

To better understand which input features contributed most to the model's ability to anticipate early warnings, a feature importance analysis was conducted on the trained Random Forest classifiers. This analysis ranks the influence of individual statistical features derived from the sensor signals. The top-ranked features are considered the most relevant for making accurate predictions.

\begin{figure}[htbp]
    \centering
    \includegraphics[width=0.81\textwidth]{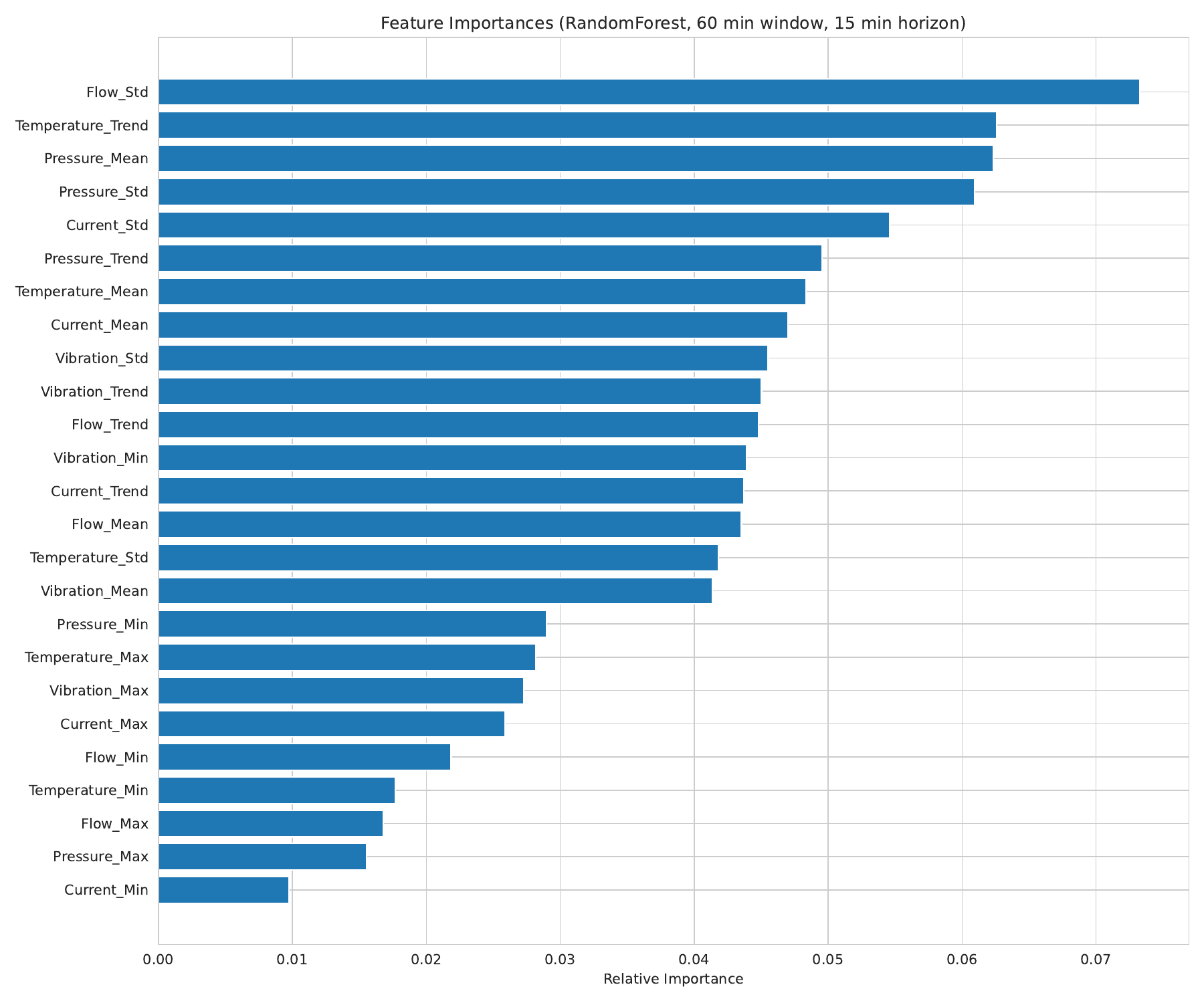}
    \includegraphics[width=0.81\textwidth]{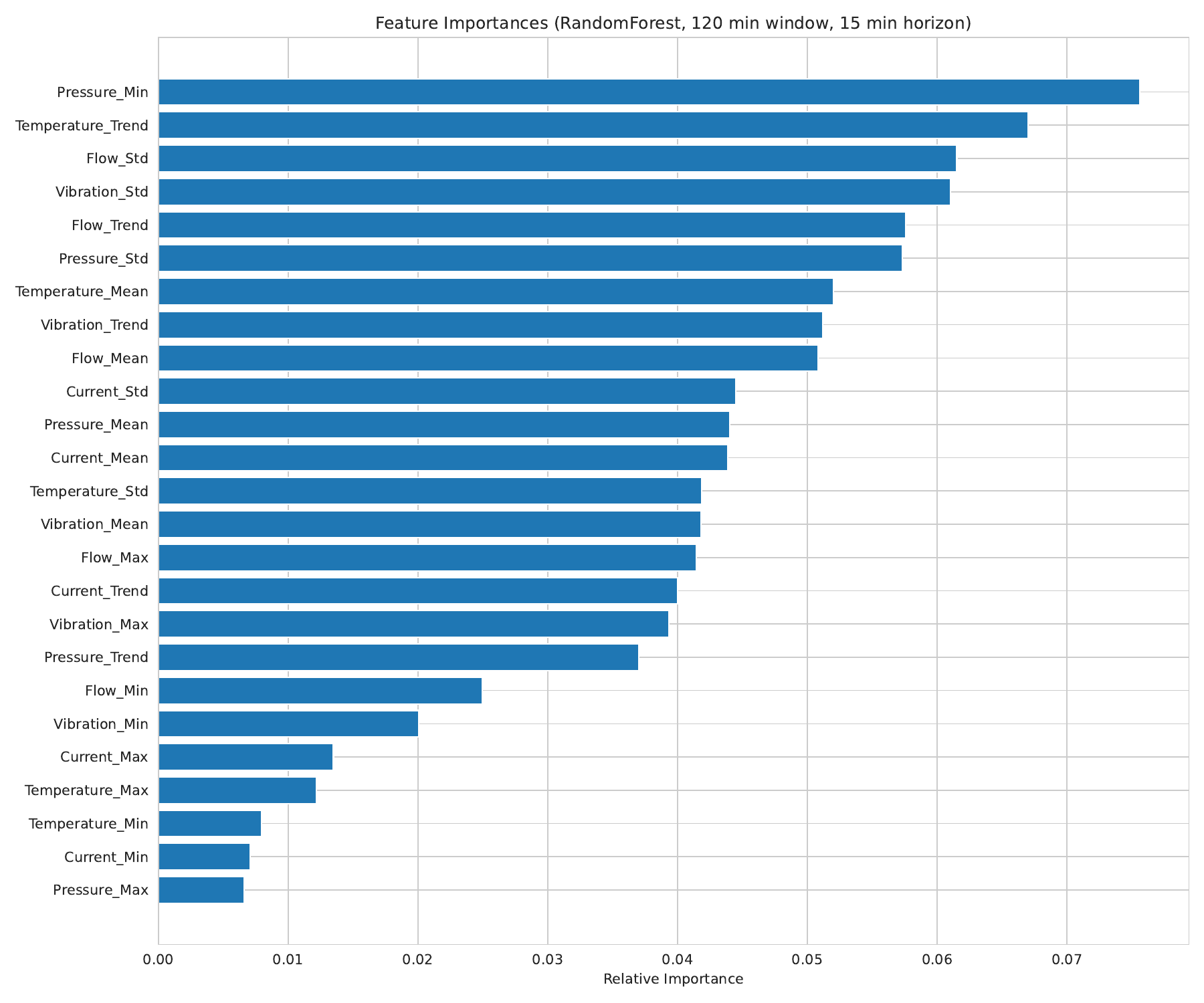}
    \caption{Feature importance rankings for the Random Forest classifier using 60-minute (top) and 120-minute (bottom) input windows. Higher values indicate greater contribution to model decisions.}
    \label{fig:feature_importance}
\end{figure}

Each input to the model consisted of multiple features extracted from the time window, including statistical summaries such as mean, standard deviation, minimum, maximum, and linear trend, calculated separately for each sensor signal. The two charts in Figure~\ref{fig:feature_importance} show the ranked feature importance for models trained with 60-minute and 120-minute input windows, respectively.

In the 60-minute model, the most important predictor was the standard deviation of the flow rate, followed by the temperature trend and pressure mean. These top features indicate that short-term fluctuations in flow and temperature trends are strong indicators of approaching abnormal behavior. Several other influential features included the standard deviation and trend of vibration and pressure signals, reflecting the system’s mechanical and hydraulic dynamics. Notably, most of the highly ranked features in the 60-minute model came from flow, temperature, and pressure signals, rather than vibration or current.

In the 120-minute model, the importance shifted toward features that reflect longer-term behavior. The minimum pressure value emerged as the most critical feature, followed closely by the temperature trend and flow standard deviation. Vibration standard deviation and flow trend also appeared among the top features. Compared to the 60-minute model, the 120-minute configuration gave more weight to features associated with baseline shifts or cumulative effects, such as minimum values and extended slopes over time. This pattern aligns with the goal of predicting events that may develop gradually over longer intervals.

Overall, the models relied on a diverse set of sensor-derived features. Flow and pressure played a consistently dominant role, particularly through their variability and long-term shifts. Vibration and current features contributed meaningfully, though typically ranked lower. These results confirm that predictive maintenance models benefit from capturing both rapid fluctuations and longer-term trends across multiple sensor dimensions.


\section{Discussion}

This study focused on short-horizon prediction of pump faults using real-time sensor data and machine learning. Instead of classifying only the present condition, the framework forecasted transitions into \textit{EarlyWarning} states at horizons of 5, 15, and 30 minutes. Such near-term prediction is valuable because it provides actionable lead time for operators, bridging the gap between long-horizon prognostics and reactive detection. Recent industrial applications confirm the importance of short-range fault forecasting in avoiding sudden shutdowns and improving operational continuity \cite{ref-cassano2025,ref-pinciroli2024,Zhang2024Electronics}.  

The dual-threshold labeling method combined fixed safety limits with adaptive statistical thresholds. This combination ensures that both absolute engineering constraints and abnormal deviations from normal patterns are captured. Similar hybrid thresholding strategies have been successfully used in auxiliary power equipment and rotating machinery under varying loads \cite{ref-li2023,ref-zonta2020}. Reviews on predictive maintenance also emphasize that combining fixed and adaptive rules improves sensitivity while maintaining interpretability for engineers \cite{ref-achouch2022}.  

Sliding window feature extraction captured temporal dynamics that precede anomalies. Results showed that shorter 60-minute windows were more effective for immediate horizons, while longer 120-minute windows provided advantages at extended horizons. This observation is consistent with prior studies, where the length of historical context was shown to influence sensitivity to fast-developing versus slowly evolving failures \cite{ref-mushtaq2021}. Feature importance analysis highlighted the strong predictive role of flow, pressure, and temperature variables, which agrees with recent findings in pump and compressor predictive maintenance \cite{ref-aminzadeh2025,ref-meitz2025}.  

The imbalance in the dataset posed a challenge, as normal cases were far more common than fault conditions. To mitigate this, SMOTE was applied to generate synthetic samples of minority classes. Imbalance is widely recognized as a limiting factor in predictive maintenance datasets \cite{Zhu2022KBS,Jiang2019IEEEAccess}. The application of SMOTE in this study is consistent with recent evidence showing that resampling improves recall and overall robustness of predictive models \cite{ref-mahale2025,ref-varalakshmi2025}.  

From a broader perspective, the framework supports the ongoing shift toward near-real-time predictive maintenance strategies. Digital twin environments increasingly require short-horizon forecasting to update maintenance schedules dynamically and optimize operation \cite{ref-aivaliotis2019}. The method presented here is interpretable and scalable, making it suitable for integration with such cyber-physical systems. Looking ahead, incorporating uncertainty estimation and combining physics-based models with machine learning may further strengthen robustness and operator trust, in line with recent advances in physics-informed predictive approaches \cite{ref-ni2023}.  

Future extensions could include multiclass prediction to differentiate fault types, online learning to adapt models to changing operating conditions, and ensemble strategies for greater resilience. Ensemble-based approaches have already demonstrated advantages in predictive maintenance for industrial assets \cite{ref-almazrouei2024}. These directions align with current trends that emphasize adaptive, interpretable, and reliable solutions for industrial predictive maintenance \cite{ref-garcia2025}.

\vspace{6pt} 





\section*{Author Contributions}
Conceptualization, K.M.A.A. and A.G.; Methodology, K.M.A.A. and A.G.; 
Software, K.M.A.A.; Validation, K.M.A.A. and A.G.; 
Formal Analysis, K.M.A.A.; Investigation, K.M.A.A. and A.G.; 
Resources, K.M.A., A.L.A.A.J., and H.A.I.A.-J.; 
Data Curation, K.M.A.A., A.G., K.M.A., A.L.A.A.J., and H.A.I.A.-J.; 
Writing—Original Draft Preparation, K.M.A.A.; 
Writing—Review \& Editing, K.M.A.A. and A.G.; 
Visualization, K.M.A.A.; Supervision, K.M.A.A. and A.G.; 
Project Administration, K.M.A.A. and A.G.

\section*{Funding}
This research received no external funding.

\section*{Institutional Review}
Not applicable.

\section*{Informed Consent}
Not applicable.

\section*{Data Availability}
The data that support the findings of this study were obtained from an industrial partner but restrictions apply to their availability. The data are not publicly available due to confidentiality agreements. They may, however, be available from the authors upon reasonable request and with permission from the industrial partner.

\section*{Acknowledgments} 
The authors acknowledge the industrial partner for providing access to sensor data and contextual information. Support from the University of Doha for Science and Technology and Humboldt-Universität zu Berlin is also acknowledged. During manuscript preparation, large-language-model tools were used for language refinement, and the authors remain fully responsible for the final content.


\section*{Abbreviations}

The following abbreviations are used in this manuscript:

\begin{center}
\begin{tabular}{ll}
ML     & Machine Learning \\
RF     & Random Forest \\
XGBoost & Extreme Gradient Boosting \\
PI     & Plant Information (Data Source) \\
RMS    & Root Mean Square \\
EWS    & Early Warning Signal \\
CA     & Critical Alert \\
SMOTE  & Synthetic Minority Oversampling Technique \\
\end{tabular}
\end{center}

\end{document}